\documentclass[11pt]{article}
\usepackage[utf8]{inputenc}
\usepackage{amsmath,amssymb}
\usepackage{graphicx}
\usepackage{booktabs}
\usepackage{authblk}
\usepackage[margin=1in]{geometry}
\usepackage{xcolor}
\usepackage[numbers,square]{natbib}
\usepackage[colorlinks=true,allcolors=blue]{hyperref}
\usepackage{caption}
\usepackage{float}
\title{\textbf{At-Grok Is Not Converged:\\ A Measurement-Validity Audit for Grokking Representation Metrics}}
\author[1]{Truong Xuan Khanh}
\affil[1]{H\&K Research Studio / Clevix LLC, Hanoi, Vietnam}
\affil[ ]{\texttt{khanh@clevix.vn}}
\date{}

\begin{document}
\setlength{\emergencystretch}{2em}
\maketitle

\begin{abstract}
On modular arithmetic, a network's embedding keeps compressing for tens of thousands of optimization
steps after it has already generalized --- so a value read at the grokking transition overstates the
converged circuit by $3$--$5\times$ and, on the modular MLP, inverts the very trend it is used to
report. A growing body of grokking research nonetheless reads representation structure from a single
such snapshot and treats it as a property of the converged circuit. We show this is unsafe in two
distinct ways, and supply the tooling to avoid it. First, a \emph{measurement hazard}: the value of a
standard representation metric --- embedding effective rank --- read at the transition can be a
transient that long training drives down to a low-rank floor. The at-grok value overstates the
converged complexity (still $1.3$--$1.5\times$ even on a transformer trained to convergence) and, on the
MLP, erases the one qualitative distinction that matters: which cells compress at all. Contra readings
in which low-rank compression coincides with the accuracy transition, compression lags it by a large
amount (of order $T_{\mathrm{grok}}$, $\ge 10^4$ steps). Second, a positive finding about \emph{what
controls} that lag: in a one-variable ablation that adds LayerNorm and nothing else, the fraction of
rank compression already completed by the grok step moves from $0.87$ to $0.25$, enlarging the
post-grok lag --- so a post-grok separation between generalization onset and representation compression
is present on every architecture we test, and its size is modulated by the normalization scheme. We
package this as a measurement-validity audit that separates onset ($T_{\mathrm{grok}}$) from compression
($T_{\mathrm{compress}}$), flags censoring, gates boundary cells that did not fully generalize, verifies
that the compression floor has itself plateaued (the same hazard can recur on the denominator), and
declines an ordering verdict when the order statistic is undefined. The audit ships with an adversarial
test suite that checks the \emph{reason} behind each verdict --- it caught a false-confidence regression
during development --- and a generic adapter that audits externally-logged runs without retraining. The
transient and the lag both recur on every architecture we test; a secondary, MLP-specific ``depth law''
relating the norm budget to the converged floor (Spearman $-1.0$) fails a pre-registered generality test
(transformer $+0.14$; protocol-reversed under free weight decay), and we report it as a negative result
on generality (Appendix~\ref{app:depthlaw}).
\end{abstract}

\section{Introduction}
Grokking --- delayed generalization long after a network fits its training set
\citep{power2022grokking} --- is increasingly studied through the \emph{representation}: the Fourier
structure of embeddings \citep{nanda2023progress}, effective rank or spectral-entropy complexity
\citep{demoss2024complexity}, intrinsic dimension, and persistent homology \citep{tang2026topological}.
A recurring move in this literature is to take a measurement at or near the grokking step --- the
moment test accuracy rises --- and read it as a property of the generalizing circuit.

This move has a failure mode that is easy to hit and hard to notice: the value of a representation
metric at the grokking step need not be its converged value, and reading it can reverse the dependence
one set out to characterize. On modular arithmetic we find a sharp instance. At grokking the embedding
effective rank is uniformly high and barely separates across budgets at the low-budget end. The
non-compressing boundary cell ($\rho=1.00$, which never compresses and stays at rank $\approx 42$) and
its strongly-compressing neighbour ($\rho=1.05$, which converges to rank $\approx 12$) read almost
identically at grok ($\approx 46$ vs.\ $\approx 48$). Stopping there, the natural reading is a high,
flat ``norm-controlled circuit complexity,'' which is exactly backwards: training each fully-generalizing
budget long past the transition drives its rank down to a low-rank floor, while the boundary cell stays
high. The at-grok snapshot overstates the converged complexity by $\approx 3$--$5\times$ and hides the
one distinction that matters (which cells compress at all).

\paragraph{Two results, kept separate.} We are careful to distinguish two claims that are easy to
conflate, because they have different epistemic status and different targets. (i) \emph{A measurement
hazard (overstatement).} Reading a representation metric at $T_{\mathrm{grok}}$ can report a value
several-fold above the converged one, and on the MLP can erase the compress-vs-not distinction. We do
not claim that careful prior work commits this error: the complexity, intrinsic-dimension, topological,
and ``construct-then-compress'' accounts track the full trajectory and document the rise and fall
directly \citep{demoss2024complexity,tang2026topological}; they are not the target. The hazard is for
the increasingly common shortcut of reading a single transition-time snapshot of a metric as if it
described the converged circuit, and our contribution against it is a tested procedure that decides when
such a reading can be trusted. The underlying practice is widespread: representation-complexity
metrics --- effective rank, intrinsic dimension, manifold geometry --- are routinely used to characterize
the generalizing solution and to compare it across a control parameter such as weight decay or task
difficulty \citep{demoss2024complexity,tang2026topological,yunis2024spectral,brown2022intrinsic}. The
hazard is specific to the version of this practice that reads the value at a single transition-time
checkpoint rather than tracking it to convergence. A representative such study would sweep weight decay
on a modular task, take each run's grokking step as the first step test accuracy crosses $0.9$, and
report the embedding effective rank (or the representation's intrinsic dimension) there as the
complexity of the generalizing solution, concluding that stronger regularization yields simpler
circuits; our audit shows this transition-time value overstates the converged complexity by
$3$--$5\times$, can invert the regularization ordering, and erases which cells compress at all
(Sec.~\ref{sec:atgrok}). We attribute no reversed conclusion to any specific published study --- the
careful work tracks the full trajectory, and one such study \citep{brown2022intrinsic} in fact reports
the same transient rise-then-fall we document, on a different metric and task --- but the snapshot
version of the practice is unguarded against the transient, and the audit is the instrument that closes
the gap. (ii) \emph{A timing result (lag).} Separately, and against a concrete reading, representation
compression does not coincide with the accuracy transition: it lags it. \citet{yunis2024spectral}
report that the validation-loss drop at grokking coincides with the simultaneous discovery of low-rank
solutions across weight matrices; at the resolution of a long-train audit we find instead that embedding
rank compression follows the accuracy transition by a large amount ($\ge 10^4$ steps, of order
$T_{\mathrm{grok}}$). This is a falsifiable correction with a named referent, and it is the one we hold
to a high bar.

\paragraph{The denominator can also mislead.} We flag at the outset a self-referential hazard the audit
must, and does, guard against: the ``converged floor'' against which we judge the at-grok value can
itself be transient. On the lowest-budget MLP cells the embedding plateaus after grokking and then, well
after it, undergoes a second collapse as the global weight norm roughly doubles
(Sec.~\ref{sec:modulates}, Fig.~\ref{fig:harness}C). A floor read over the final tenth of training would
sit inside this second regime. The same discipline we apply to the numerator therefore applies to the
denominator: a careful audit verifies that the floor has plateaued, not merely that training has run
long. We treat this not as a caveat to the thesis but as a sharpening of it.

\paragraph{What modulates the lag.} Beyond establishing the lag, we identify what controls its size, and
this is the paper's main positive finding. Across a single harness that changes one architectural
variable at a time (Sec.~\ref{sec:modulates}), the fraction of embedding-rank compression already
completed by the grok step --- \emph{frac-pre} --- moves from $0.87$ on a canonical attention model to
$0.25$ when LayerNorm is added and nothing else, with the MLP intermediate at $0.66$. A high frac-pre
leaves a small residual lag; a low one produces a large one --- but a non-zero lag is present even at
the highest frac-pre we observe (the canonical transformer still settles within $\approx 0.63\,
T_{\mathrm{grok}}$ of grokking, Sec.~\ref{sec:transformer}). The grok-to-compression separation is
therefore general, and its magnitude is modulated by the normalization scheme --- a result that recasts
``how separated are onset and compression?'' as a question about architecture rather than about the
phenomenon.

\paragraph{Contributions.} Our findings sit at two confidence tiers, kept distinct throughout. The
central results are well supported and hold across every architecture we tested: the at-grok transient,
the grok-to-compression lag, and the normalization-mediated control of that lag. A secondary observation
is deliberately narrower --- an MLP-specific depth law that our own pre-registered generality test
falsifies --- and we say which is which at every step.
\begin{itemize}
\item \textbf{An at-grok-vs-converged audit (the measurement hazard).} We show, on modular addition and
multiplication MLPs, that embedding effective rank at grokking is a transient peak that collapses under
long training; the converged solution is low-rank at every fully-generalized norm budget (the one budget
that only partially generalizes is the boundary case that does not compress, and the audit gates it
out). The transient is not an artifact of the clamp intervention or of arithmetic: it recurs under a
free weight-decay protocol and on parity, with the at-grok rank exceeding the converged floor in every
generalizing cell across three tasks and both protocols, and it is metric-agnostic (it appears in
participation ratio and stable rank as well).
\item \textbf{A compression-lag diagnostic.} We define $T_{\mathrm{grok}}$, $T_{\mathrm{compress}}$,
their lag, a censoring flag, a boundary gate that excludes cells that did not fully generalize, and a
floor-plateau check. The audit reports a large separation of the two times for both tasks, with
generalization preceding compression by a lag of order $T_{\mathrm{grok}}$ ($\ge 10^4$ steps), while
explicitly declining a single-clock claim on the enlarged grid.
\item \textbf{What modulates the lag: a post-grok lag is general, and normalization enlarges it (a
positive result).} In a one-variable ablation (MLP vs.\ a LayerNorm-free canonical transformer vs.\ that
transformer with LayerNorm added), the single LayerNorm change moves frac-pre --- the fraction of the
total embedding-rank compression already completed by the grok step --- from $0.87$ to $0.25$ and
enlarges the post-grok lag. A lag is present on every arm; LayerNorm makes it large. Within this harness
it is LayerNorm, not ``being a transformer,'' that defers compression past grokking; the large lag the
MLP shows is normalization-mediated rather than a property of grokking. We scope this to the power we
have (see below).
\item \textbf{A test suite that enforces honesty.} An adversarial synthetic suite checks that the audit
does not fabricate compression times under censoring, high-floor boundary cells, rebound, or
compression-before-grok, and that a clock-type verdict is backed by a finite order statistic rather than
a matching string. This caught a real regression during development.
\item \textbf{A transformer dose-response: the transient generalizes.} We run a well-powered transformer
clamp+free sweep trained to convergence (Sec.~\ref{sec:transformer}). It generalizes the central claim
--- the at-grok transient recurs on the embedding and unembedding (median $1.3$--$1.5\times$), is
metric-agnostic, and is present at every weight and activation locus --- and the lag recurs. It also
delivers a clean negative result on the secondary depth law (Appendix~\ref{app:depthlaw}).
\item \textbf{Scope.} We do not re-claim compression in grokking (prior work establishes it under
several lenses); the contribution is a tested protocol for deciding whether a given representation
metric has converged, is censored, or remains transient at $T_{\mathrm{grok}}$, the norm-budget
quantification of the grok-to-compression lag, and the identification of normalization as the variable
that sets it.
\end{itemize}

\paragraph{What the audit checks, at a glance.} Given per-step logs of a representation metric and test
accuracy across norm budgets, the audit: (i) separates the onset clock $T_{\mathrm{grok}}$ (first step
at the grokking accuracy threshold) from the compression clock $T_{\mathrm{compress}}$ (first post-onset
step at which the metric settles within a tolerance of its floor); (ii) gates boundary cells that never
fully generalize or never compress before computing any ordering statistic; (iii) verifies the floor
itself has plateaued, since the same transient hazard can recur on the denominator; (iv) returns
low-power / descriptive only when fewer than two non-boundary cells remain; and (v) declines a
clock-type verdict whenever the cross-budget order statistic is undefined. A nine-case adversarial suite
fixes each verdict and its reason in advance.

\section{Related Work}
We group prior work by what it measures and where our audit sits relative to it
(Table~\ref{tab:positioning}). The recurring theme is that several lines have established that grokking
involves compression or geometric reorganization; none separates onset from compression with censoring
and boundary handling, quantifies the lag as a norm-budget dose-response, or ships a tested audit.

\paragraph{Complexity, rank, and compression in grokking.} Several lines establish that grokking
involves a fall in representational complexity. \citet{demoss2024complexity} introduce a Kolmogorov/MDL
complexity (lossy compression of coarse-grained weights) that rises during memorization and falls at
generalization. Most directly related to us, \citet{yunis2024spectral} observe a task-agnostic view in
which the validation-loss drop at grokking coincides with the simultaneous discovery of low-rank
solutions across all weight matrices. Our central empirical correction is precisely to this
``coincides'': at the resolution of a long-train audit, embedding rank compression does not coincide
with the accuracy transition but lags it by a large, norm-budget-dependent amount ($\sim T_{\mathrm{grok}}$).
\citet{yunis2024spectral} also note, qualitatively, that under higher weight decay the smaller singular
values disappear while under none they do not. On the modular-MLP clamp we observe a clean monotone
counterpart, the converged effective-rank floor falling with the norm budget (Spearman $-1.0$), but we do
not claim it as the quantitative form of their general observation. That clean law is protocol- and
architecture-specific: our own free weight-decay sweep does not reproduce it and a transformer clamp
sweep shows no ordering at all (Sec.~\ref{sec:transformer}, Appendix~\ref{app:depthlaw}). What we share
with \citet{yunis2024spectral} is the part that is robust --- that low-rank structure accompanies
generalization --- not a general norm law. For the same reason, the apparent tension with
\citet{alquabeh2025embedding}, who track the rank of the embedding $E$, the first layer $W$, and their
product under small weight decay ($0.001$--$0.005$) and report that $E$'s rank stays largely stable while
$W$ and $EW$ compress, is an open cross-setup question that our clamp law does not resolve, and we are
explicit that it does not. Their weak weight decay corresponds, at equilibrium, to a large embedding norm
(their own steady-state analysis gives $\lVert e_i\rVert \propto 1/\lambda$), i.e.\ the high-budget end of
our norm axis, where the clamp law would if anything predict strong embedding compression --- the opposite
of their stable $E$. Their setting is moreover a free-decay one, where our own data shows the clean
monotone floor law does not hold (Appendix~\ref{app:depthlaw}). We therefore do not present the clamp law
as explaining their stable-$E$ observation; a clean reconciliation would require the same protocol,
prime, and architecture on both sides --- the kind of same-run comparison our metric-agnostic clock is
built for. A related observation appears in \citet{wang2026dimensional}, whose primary object is an
effective dimensionality of gradient-avalanche dynamics (not a weight rank); as a secondary structural
signature they report a transient peak in the Gini concentration of the full parameter vector that
coincides with grokking. Our quantity is different (the effective rank of a single weight matrix, the
embedding) and so is our finding: its compressed endpoint is reached well after onset, and it is that
onset-to-floor lag, not a peak located at the transition, that we measure. A related geometric line
reaches a compatible picture from a different metric (a Jacobian-alignment score), describing a
``construct-then-compress'' dynamic in transformers with an explicit post-grokking refinement phase. The
direction in which a metric moves relative to grokking is itself metric-dependent. A concurrent example
makes this vivid: \citet{sivasankar2026circuit} show that a permutation-tested Fourier
circuit-synchronization measure reaches its post-grok level roughly $500$--$3{,}000$ steps \emph{before}
grokking on modular addition (mean lead $\approx 1{,}700$ steps) --- the opposite timing to the
embedding-rank compression we date after it. That two representational quantities move in opposite
directions across the same transition is a direct illustration of why \emph{which} quantity one reads,
and \emph{when}, determines the picture. We build on these: the novelty is not that compression occurs,
but that the common practice of reading rank at $T_{\mathrm{grok}}$ measures a transient, that the
onset-to-compression lag is itself a measurable quantity, and that normalization modulates it ---
packaged as a tested audit.

\paragraph{Topological and geometric diagnostics.} \citet{tang2026topological} give a topological
characterization of grokking, computing persistent homology on point clouds derived from the embedding
matrices --- the same locus we audit: first-homology persistence rises sharply at generalization, and
they compare persistent homology against Fourier analysis and local intrinsic dimension (LID), reporting
that LID drops (a compression onto a low-dimensional manifold) with the topological loop becoming
apparent only after that compression. Their cross-correlation analysis finds that test accuracy precedes
the topological transition by $\sim 10^3$ steps. Our rank-compression lag is $\sim 2\times10^4$ steps. If
the two diagnostics were tracking the same event under comparable conditions, one might expect their
lags to be closer; the large difference therefore motivates a same-run comparison, but does not by itself
establish distinct events, since the two numbers come from different tasks, models, and lag definitions
(a topological feature of the embedding point cloud vs.\ the effective rank of the embedding matrix). We
state the distinctness only as a hypothesis. Relative to this line, our metric is different (effective
rank vs.\ PH/LID), our question is different (is a transition-time snapshot converged?), and our
contribution is a tool, not a new geometric signature.

\paragraph{Transition-localization diagnostics.} A 2026 line treats grokking detection itself as a
diagnostic problem, but with a different target than ours: predicting or localizing the transition,
typically before accuracy rises. \citet{wang2026spectral} map trajectory observables to
Wasserstein/quantile coordinates and a Hankel-DMD residual (alongside spectrum and effective rank) and
report run-level grok-vs-no-grok discrimination with a threshold/false-alarm/lead-time trade-off;
\citet{verma2026weightdecay} gives cheap online weight-decay-regime diagnostics on grokking transformers;
and \citet{xu2026earlywarning} uses a loss-landscape commutator defect as an early-warning signal with a
power-law lead time. \citet{golwala2026ildr} proposes a held-out inter/intra-class distance ratio as an
early-detection signal and, in passing, reports independently that the weight norm consistently lags the
transition --- the same post-grok lag we measure, read there from parameter space rather than from the
representation. Relatedly, \citet{xu2026multitask} sweeps weight decay as a phase parameter and finds the
grokking timescale and curvature covary with it; they also report a ``holographic incompressibility'' in
which the converged multi-task solution, though confined to a few principal trajectory directions,
remains algebraically full-rank and resists post-hoc SVD truncation. This is compatible with our finding
rather than in tension with it: a low effective rank (a concentrated singular-value spectrum) does not
imply exact rank-deficiency, and the small singular directions Xu shows to be load-bearing are exactly
what an effective-rank floor leaves in place. Our norm-budget axis is the same control knob they sweep,
but we use it to date a post-onset compression event rather than to characterize the converged geometry.
These are forward (or geometry-characterizing) analyses operating at or before onset, or on the converged
solution's structure. Ours is a backward audit (was this post-onset measurement converged?): it operates
after $T_{\mathrm{grok}}$, on the question of whether a representation metric read at the transition
reflects the converged circuit. The two are complementary: a forward detector says a transition is
coming, and our audit says whether a structural reading taken at it can be trusted. The premise that a
single transition-time snapshot under-determines the converged network is, in fact, supported from
several independent directions. \citet{prakash2026antigrokking} show, by training far past
generalization, a late-stage ``anti-grokking'' phase in which test accuracy can collapse back to chance
while training accuracy stays perfect --- so a reading taken at $T_{\mathrm{grok}}$ need not even survive
to late training. \citet{xu2026multitask} show, from the opposite side, that a converged solution can
look low-dimensional along its trajectory yet remain load-bearing in its small singular directions, so a
truncated reading of it fails. Our contribution is to make this premise operational for the common
practice of reading effective rank: rather than warning in prose that snapshots can mislead, we give a
clock that dates when the metric actually settles, flags when a run is too short to tell, and ships the
check as a tested tool.

\paragraph{A measurement-reliability strand.} Most grokking research asks what the phenomenon is or what
produces it. A smaller strand attends instead to \emph{how} grokking is measured: whether the quantities
and thresholds used to study it are reliable, and how one would check. \citet{miller2024sharpness} make
this concern explicit for the transition itself: they fit a shared functional form to the train- and
test-accuracy curves so that ``sharpness'' and jump-time are extracted in a noise-robust,
assumption-checkable way rather than read off raw curves, and they note the framework could be used to
test consistency across existing works. \citet{prakash2026antigrokking} make the complementary move on
structural diagnostics, comparing which metrics ($\ell_2$ norm, activation sparsity, weight entropy,
spectral heavy-tailedness) remain faithful into a late anti-grokking phase and which silently fail. Our
audit shares this concern but turns it on a specific, widely-used measurement --- the effective rank read
at the transition --- and contributes the missing piece for it: not just a caution that the reading can
mislead, but a tested procedure that decides when it can be trusted. Independent support that this
reading is unsafe comes from \citet{brown2022intrinsic}: tracking the last-layer intrinsic dimension of
activations across grokking, they observe a transient rise at the onset of generalization followed by a
descent to a lower floor --- a transition-time peak in a representation-complexity measure of precisely
the kind a single-snapshot reading would mistake for the converged value.

\paragraph{Relation to the weight-norm line.} The norm-clamp protocol places this audit in the
weight-norm line of grokking work, which we delineate precisely. \citet{liu2023omnigrok} establish that
the weight norm governs grokking --- grokking is the drift from a high initialization norm to a
generalizing one --- and a body of work characterizes how the norm and weight decay set \emph{when}
grokking happens: weight-decay phase structure and timescales \citep{xu2026multitask,verma2026weightdecay},
circuit-efficiency pressure toward the generalizing solution \citep{varma2023circuit}, and the
lazy-to-rich transition the norm induces \citep{kumar2024lazy}. That line explains \emph{when} grokking
happens as a function of norm; the present work asks the orthogonal question of whether a representation
metric read at the transition has converged, and uses the same norm knob only to date a post-onset
compression event. This line also connects to our own weight-norm delay-law study
\citep{truong2026normsep}, which establishes that the grokking delay itself scales with the norm ratio
between memorizing and structured solutions; that result is about \emph{when} grokking happens, whereas
the present audit is about whether a metric read \emph{at} that moment has converged, and the two are
complementary rather than overlapping. Mechanistic interpretability work that reverse-engineers what
grokking networks compute --- Fourier-feature circuits and restricted-loss progress measures
\citep{nanda2023progress}, and analyses of the memorizing-vs-generalizing circuit competition --- measures
structure that is meaningful at and around the transition; our audit is complementary and prior to it,
providing a check on when a transition-time measurement of any such metric describes the converged
circuit rather than a passing state.

\begin{table}[H]
\centering
\footnotesize
\setlength{\tabcolsep}{4pt}
\caption{Where this audit sits relative to prior grokking-dynamics work. Prior lines establish that
compression/reorganization happens (some asserting it coincides with grokking); none separates onset
from compression with censoring and boundary handling, quantifies the lag as a norm-budget
dose-response, or ships a tested audit. ``Lag vs.\ ctrl'' = does the work measure the onset-to-compression
lag as a function of a control parameter. ``Dir.'' (direction) = predict the transition (forward) vs.\
audit a post-transition measurement (backward).}
\label{tab:positioning}
\begin{tabular}{@{}p{0.215\textwidth}p{0.135\textwidth}cccccc@{}}
\toprule
 & metric & onset/ & lag/ & bdry/ & tested & dir. \\
 & & compr. & ctrl & cens. & audit & \\
\midrule
Yunis et al.\ '24 (spectral dyn.) & weight rank & \emph{coinc.} & no & no & no & --- \\
DeMoss et al.\ '25 (complexity) & MDL/Kolm. & qual. & no & no & no & --- \\
Tang et al.\ '26 (topology) & PH, LID & CCF & no & no & no & --- \\
Wang et al.\ '26 (spectral diag.) & DMD res., e.r. & n/a & no & no & part. & fwd \\
\textbf{This work} & effective rank & \textbf{lag} & \textbf{yes} & \textbf{yes} & \textbf{yes} & bwd \\
\bottomrule
\end{tabular}
\end{table}

\section{Setup and definitions}
\label{sec:setup}
\paragraph{Task and model.} We study modular addition and multiplication mod $p$ ($p=59$) with a
two-layer MLP over learned token embeddings ($d=128$, hidden $H=256$, GELU), trained full-batch with
AdamW and decoupled weight decay. This is the standard grokking setting; our analysis is about when
representation structure is measured, not about the architecture.

\paragraph{Norm budget.} The weight norm governs grokking: \citet{liu2023omnigrok} show that grokking is
a slow drift from a large initialization norm toward a generalizing norm, so that setting the norm
directly can speed up or remove the delay. We use this knob in a matched form, holding the global
parameter norm at $\rho\lVert W\rVert_c$ throughout training, where $\lVert W\rVert_c$ is the norm the
free dynamics reach at grokking. Sweeping $\rho$ gives a family of runs that grok at different speeds; we
ask how the post-grok representation evolves in each. The clamp is fully specified here and used only as
a standard control knob, not as a contribution.

\paragraph{Effective rank.} For a weight matrix $M$ with singular values $\sigma_i$ we report the
variance-normalized spectral-entropy effective rank $\exp\!\big(-\sum_i \bar\sigma_i\log\bar\sigma_i\big)$,
$\bar\sigma_i=\sigma_i^2/\sum_j\sigma_j^2$ --- the squared (explained-variance) normalization used in the
complexity-of-grokking work we build on \citep{demoss2024complexity}. The effective-rank construction is
due to \citet{roy2007effective}, who normalize the singular values directly ($\bar\sigma_i =
\sigma_i/\sum_j\sigma_j$); the form we use squares them (equivalently it is the spectral entropy of the
eigenvalues of $M^\top M$). Both are maximized by a flat spectrum and equal one at rank one, so they
capture the same notion of spectral spread; we use the squared form for continuity with prior grokking
work and note the variant only for exact reproducibility. The choice of a singular-value--based quantity
is principled rather than merely inherited: phase-transition accounts of grokking associate each
learnable feature with a singular value \citep{ersoy2026noise}, so the effective number of active
singular directions is a natural quantity to watch cross the transition. The metric is not novel here;
our contribution is the axis on which it is read (post-grok time and norm budget) and the discipline
around it.

\paragraph{Clocks.} We define $T_{\mathrm{grok}}$ as the first step with median test accuracy $\ge 0.9$,
and $T_{\mathrm{compress}}$ as the first step after $T_{\mathrm{grok}}$ at which the median effective
rank falls within $\varepsilon$ of its final-plateau floor ($\varepsilon=0.1$ of the at-grok-to-floor
drop). The lag is $T_{\mathrm{compress}}-T_{\mathrm{grok}}$.

\section{At-grok is not converged}
\label{sec:atgrok}
Measuring effective rank only at $T_{\mathrm{grok}}$ yields a high profile across $\rho$ that barely
separates the low-budget cells (the boundary cell and its compressing neighbours read nearly alike) and
invites a ``converged complexity'' reading. Training each budget to $2\times10^5$ steps refutes that
reading: at every budget that fully generalizes, the effective rank falls to a low-rank floor while test
accuracy reaches $1.0$, and the at-grok profile flattens at convergence. The at-grok value is a
transient. This holds on both tasks. The drop magnitude is itself budget-dependent: budgets that grok
fastest also start their compression from the highest at-grok rank and fall furthest. On modular
addition, effective rank falls from $\approx 50$ at grok to $\approx 7$ at convergence in the
fast-grokking cells, while the slowest budget still falls from $\approx 10$ to $\approx 4$. The norm
budget thus shapes the trajectory to the compressed solution rather than the converged complexity, which
is low everywhere generalization completes.

\begin{figure}[H]
\centering
\includegraphics[width=\textwidth]{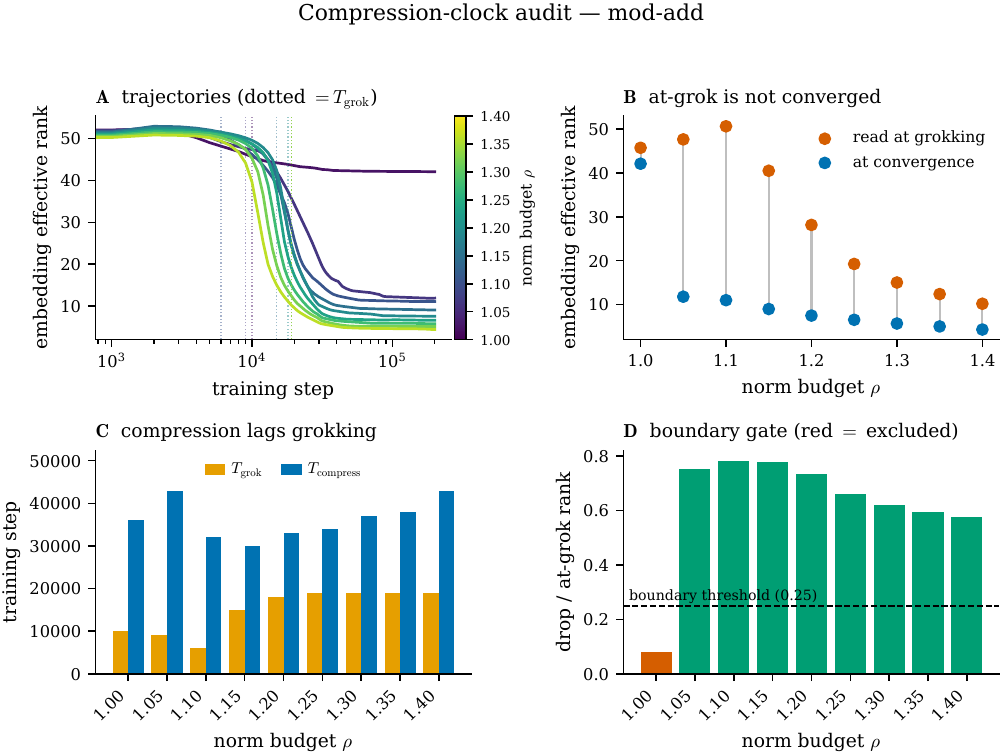}
\caption{Compression-lag dashboard on the modular-addition MLP ($p=59$, 6 seeds, real data), the primary
result. \textbf{A}: embedding effective-rank trajectories per norm budget $\rho$ (dotted: $T_{\mathrm{grok}}$);
fast-grokking budgets fall from $\approx 50$ to a low floor, while the floor cell $\rho=1.00$ (dark)
stays near $42$. \textbf{B}: effective rank at grokking (red) vs.\ at convergence (blue) --- the at-grok
value sits well above the converged floor at every budget that completes, i.e.\ it is a transient; only
at $\rho=1.00$ do the two coincide (no compression). \textbf{C}: $T_{\mathrm{grok}}$ (orange) vs.\
$T_{\mathrm{compress}}$ (teal); compression lags grokking at every budget. \textbf{D}: the boundary
gate --- the $\rho=1.00$ cell (red) has a drop/at-grok ratio below threshold and is excluded before the
ordering statistics are computed, preventing a spurious two-clock verdict.}
\label{fig:mlp}
\end{figure}

A boundary case sharpens the point. At the floor budget the network only partially generalizes (accuracy
plateaus near $0.99$, never reaching $1.0$) and the embedding does not compress: it remains high-rank
through $2\times10^5$ steps. This is a capacity boundary, not a second behavior of the clock, and (as we
discuss next) a naive clock analysis that does not exclude it can manufacture a spurious ``two clocks''
conclusion. Figure~\ref{fig:mlp} shows the full picture for modular addition; modular multiplication
behaves identically, with every swept budget compressing and no floor cell in range.

\begin{figure}[H]
\centering
\includegraphics[width=\textwidth]{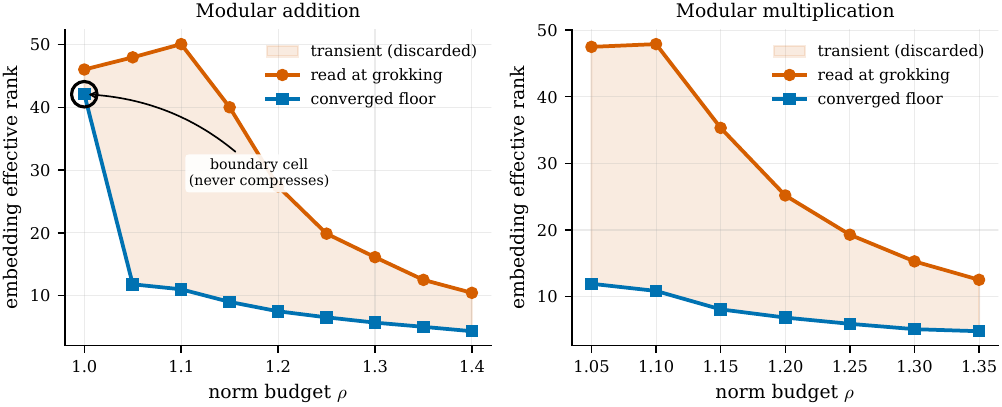}
\caption{Reading effective rank at grokking inverts the picture. Embedding effective rank read at
grokking (red) vs.\ the converged floor after long training (blue), against the norm budget $\rho$; the
grey band is the transient the converged network discards. The converged curve is the monotone depth
law; the at-grok curve overstates it by $3$--$5\times$ and is nearly flat where the converged curve falls
steepest. On addition (left) the boundary cell $\rho=1.00$ (circled) reads the same as its neighbours at
grok but never compresses --- a distinction the snapshot erases and only long training reveals.}
\label{fig:atgrok}
\end{figure}

\paragraph{The concrete failure mode.} Figure~\ref{fig:atgrok} makes the hazard explicit by plotting the
effective rank read at grokking against the converged floor, both as functions of the norm budget. The
two curves tell qualitatively different stories. The converged curve is the depth law: a strict monotone
fall from the boundary cell down to a low floor. The at-grok curve is high and nearly flat at the
low-budget end, then falls, and at every budget that compresses it sits $3$--$5\times$ above the
converged value (median $3.3\times$). Most sharply, the at-grok snapshot cannot separate the one cell
that does not compress from the ones that do: the boundary cell ($\rho=1.00$, converged rank $\approx
42$) and its neighbour ($\rho=1.05$, converged rank $\approx 12$) read $46$ and $48$ at grok, a $4\%$
difference that hides a $3.6\times$ difference in the converged solution. A study that read embedding
rank at the transition would therefore report a circuit complexity several-fold too high, miss that one
budget never compresses, and recover a complexity-vs-budget shape that the converged network does not
have. This is the failure mode the audit exists to catch; it is not hypothetical, and it is invisible
without training past the transition.

\paragraph{Not specific to the clamp protocol or to modular arithmetic.} The transient is the paper's
central claim, and it depends on neither the norm-clamp intervention nor modular arithmetic. On a free
weight-decay sweep at $p=59$, with no norm clamp at all, the at-grok embedding rank exceeds the converged
floor in every fully-generalizing cell on modular addition (median $37\rightarrow 18$), modular
multiplication ($35\rightarrow 18$), and parity, a non-arithmetic task ($23\rightarrow 3$): a
$2$--$7\times$ overstatement, with parity showing the largest transient
(\texttt{transient\_replication.py}). The transient is also metric-agnostic. On the same free-decay
addition cells the at-grok overstatement appears in participation ratio ($2.2\times$) and stable rank
($1.4$--$1.7\times$) just as in effective rank ($2.5$--$2.6\times$), and the one decay setting with no
effective-rank transient shows none in the other two either, so the hazard is a property of the
representation's spectral compression rather than of the particular rank measure
(\texttt{metric\_agnostic\_transient.py}). Independently, and on a different task and a different metric,
\citet{brown2022intrinsic} report the same qualitative signature while tracking the last-layer intrinsic
dimension of activations across grokking: on their slower-grokking modular-division runs the intrinsic
dimension shows a transient rise at the onset of generalization, followed by a descent to a value below
the earlier plateau --- a transition-time peak in a representation-complexity measure, exactly the
structure a single-snapshot reading would mistake for the converged one. \citet{brown2022intrinsic} read
this spike-then-descent as an escape from a local minimum; for our purposes the point is only that the
value read at the transition is not the converged one, but the agreement across a different metric, a
different task, and an independent mechanistic interpretation is what makes the transient hard to dismiss
as an artifact of our setup. One thing does not carry over, and we flag it rather than bury it: the depth
law's direction. Under the clamp protocol the converged floor falls with the norm budget; under free
weight decay it rises with the decay strength, and the two do not collapse onto a single
floor-versus-converged-norm curve (Appendix~\ref{app:depthlaw}). The transient generalizes across
protocol and task; the depth law's sign is protocol-specific.

\section{Generalization precedes compression: a large lag}
\label{sec:lag}
\begin{figure}[H]
\centering
\includegraphics[width=\textwidth]{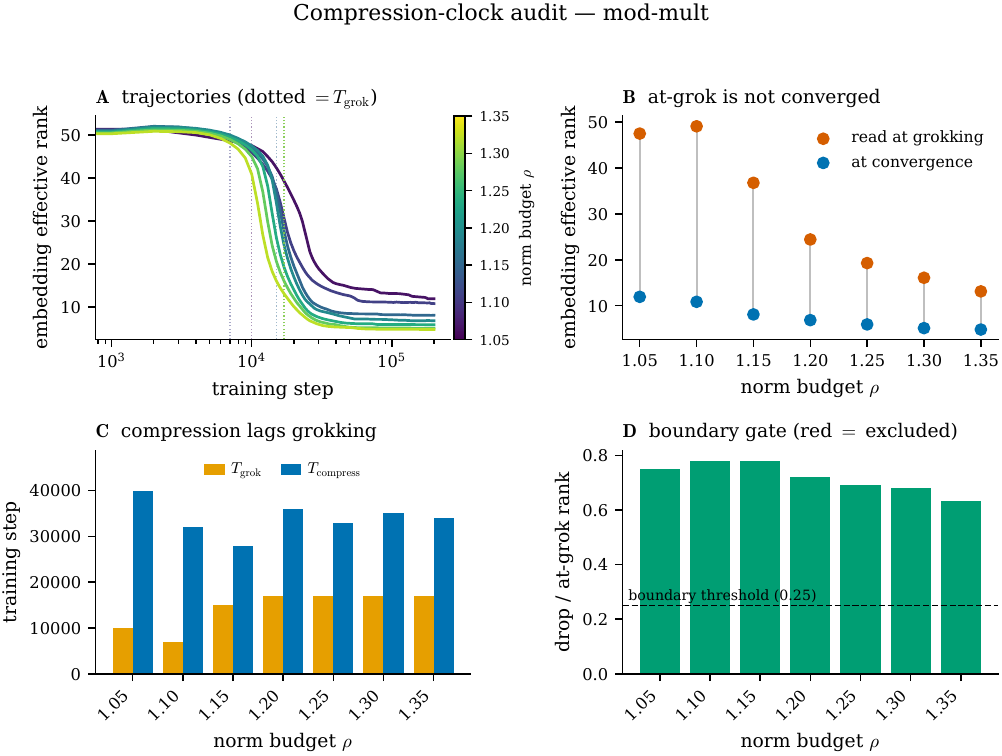}
\caption{The same audit on the modular-multiplication MLP ($p=59$, 6 seeds, real data), a replication of
Figure~\ref{fig:mlp}. The at-grok transient (B) and the compression lag (C) recur; here the swept
budgets ($\rho=1.05$--$1.35$) all lie above the functional floor, so all cells compress and none is
gated (D). The post-grok lag is comparable to $T_{\mathrm{grok}}$ on both tasks
($\mathrm{lag}/T_{\mathrm{grok}}=1.04$ here vs.\ $1.00$ on addition), illustrating that the same
qualitative separation recurs with a task-specific coefficient.}
\label{fig:mult}
\end{figure}

On both modular addition and multiplication the audit returns \textsc{partially separated clocks +
large lag} (Table~\ref{tab:clocks}). Compression follows grokking by a lag comparable to the
time-to-grok itself: median lag $1.7\times10^4$ steps on multiplication ($\mathrm{lag}/T_{\mathrm{grok}}
=1.04$) and $1.8\times10^4$ on addition ($\mathrm{lag}/T_{\mathrm{grok}}=1.00$); Figure~\ref{fig:mult} is
the multiplication dashboard. This large separation is the robust, replicated result. The norm budget
strongly orders $T_{\mathrm{grok}}$ ($\rho_S(\rho,T_{\mathrm{grok}})=+0.90,+0.85$) but does not order
$T_{\mathrm{compress}}$ ($\rho_S(\rho,T_{\mathrm{compress}})=+0.18,+0.29$): generalization onset and
representation compression are temporally separable, with a real and large gap, but the compression time
is not a clean function of the budget. We therefore report a large separation rather than a single shared
clock.

We state the robust form of the lag carefully. Its magnitude is of order $T_{\mathrm{grok}}$ and always
$\ge 10^4$ steps; the exact ratio $\mathrm{lag}/T_{\mathrm{grok}}$ depends on the compression tolerance
$\varepsilon$ (which sets how close to the floor counts as ``compressed'') and ranges over $[0.65,1.46]$
as $\varepsilon$ varies from $0.20$ to $0.05$ (Appendix~\ref{app:sensitivity}, Table~\ref{tab:sensitivity}).
We report $\varepsilon=0.10$, where $\mathrm{lag}/T_{\mathrm{grok}}\approx 1.0$, as the default, and treat
``$\ge 10^4$ steps'' rather than ``$=T_{\mathrm{grok}}$'' as the claim that does not depend on the
tolerance.

\begin{table}[H]
\centering
\small
\caption{Compression-lag audit on the two modular-arithmetic MLPs ($p=59$, 6 seeds per cell, $2\times10^5$-step
budget), on the enlarged budget grid. The large post-grok lag and the monotone floor law replicate on
both tasks; the compression time is not cleanly ordered by the budget. On modular addition the boundary
gate excludes the floor cell $\rho=1.00$, which only partially generalizes and never compresses, before
the ordering statistics are computed.}
\label{tab:clocks}
\begin{tabular}{lcc}
\toprule
 & modular multiplication & modular addition \\
\midrule
cells (norm budgets $\rho$) & 7 ($1.05$--$1.35$) & 9 ($1.00$--$1.40$) \\
median gap $T_{\mathrm{compress}}-T_{\mathrm{grok}}$ & $17{,}000$ & $18{,}000$ \\
$\mathrm{lag}/T_{\mathrm{grok}}$ & $1.04$ & $1.00$ \\
$\rho_S(\rho,T_{\mathrm{grok}})$ & $+0.90$ & $+0.85$ \\
$\rho_S(\rho,T_{\mathrm{compress}})$ & $+0.18$ & $+0.29$ \\
$\rho_S(\rho,\text{converged floor})$ & $-1.00$ & $-1.00$ \\
boundary cell excluded & --- & $\rho=1.00$ \\
verdict & part.\ separated $+$ large lag & part.\ separated $+$ large lag \\
\bottomrule
\end{tabular}
\end{table}

\paragraph{A corrected reading (the audit corrects its own authors at the data level).} An earlier,
smaller grid (four and five budget cells) showed same-signed $T_{\mathrm{compress}}$ correlations
($+0.40$ and $+0.95$) that invited a ``one clock'' reading. Enlarging the grid to seven and nine cells
weakened that ordering sharply --- the multiplication $T_{\mathrm{compress}}$ correlation fell to $+0.18$
and addition to $+0.29$, neither an ordering --- while leaving the large lag and the floor law intact. We
report this openly: it is the same standard the audit enforces against others, that a correlation on four
points is not an ordering, applied to our own earlier reading. Both readings are reproducible:
\texttt{corrected\_reading.py} recomputes $\rho_S(\rho,T_{\mathrm{compress}})$ on the small grid ($+0.95$
addition, $+0.40$ multiplication) and the full grid ($+0.29,+0.18$) from the data shipped in
\texttt{sample\_data/}.

\paragraph{Why the boundary gate matters.} If the partially-generalizing floor cell is left in, its high,
never-compressing rank produces a degenerate $T_{\mathrm{compress}}$ that, combined with the genuine
cells, can flip the global ordering and yield a false \textsc{two clocks} verdict. On modular addition
this is concrete: the $\rho=1.00$ cell sits at effective rank $42.1$ at convergence (versus a typical
floor of $7.5$ for the cells that complete), a drop of only $0.09$ relative to its at-grok value, since
it groks only partially and never compresses. The audit detects such cells by their tiny drop-to-at-grok
ratio and high floor, excludes them from the dose-response, and reports them separately as boundary
cells. This single guard is the difference between a fragile finding and a robust one.

\section{What modulates the lag: a post-grok lag is general, and normalization enlarges it}
\label{sec:modulates}
Section~\ref{sec:lag} establishes that compression lags grokking. Here we identify what controls the
size of that lag, which is the paper's main positive finding. The central claim of
Section~\ref{sec:atgrok} is established on the modular MLP and, in Section~\ref{sec:transformer}, on a
transformer under the clamp. To test whether the lag depends on being a transformer, on attention, on
normalization, or on initialization --- which differ all at once between the two --- we run a single
free-weight-decay harness that changes one thing at a time: arm A is the MLP anchor; arm B is a canonical
attention model that is LayerNorm-free (ReLU MLP block, no biases, $1/\sqrt d$ initialization); arm C is
arm B with LayerNorm added and nothing else. A-vs-B isolates architecture; B-vs-C isolates LayerNorm.
Each arm is swept under free weight decay (MLP $\{0.5,1,2,4\}$; transformer $\{1,1.5,2,3\}$), $p=59$,
train fraction $0.3$, up to $6\times10^4$ steps, every quantity computed by the same per-seed clock as
the MLP audit.

\paragraph{The transient holds on all three arms.} The at-grok transient exceeds the converged floor on
all three arms: $\approx 1.5\times$ on the MLP, $\approx 1.5\times$ on the canonical transformer, and
$\approx 3.2\times$ on the LayerNorm transformer. As on the modular MLP, reading embedding rank at the
transition overstates the converged value on every architecture we tried; the hazard is not specific to
the MLP or to the clamp.

\paragraph{What the lag depends on: how much compression precedes grokking.} The arms differ sharply in
when the embedding compresses relative to grokking. Define $\mathrm{frac\text{-}pre} = (r_{\mathrm{init}}
- r_{\mathrm{grok}})/(r_{\mathrm{init}} - r_{\mathrm{floor}})$, the fraction of the total embedding-rank
compression already completed by the grok step. It is $0.87$ on the canonical transformer (most
compression precedes grokking, leaving a small residual lag and a small transient), $0.66$ on the MLP,
and $0.25$ on the LayerNorm transformer (most compression follows grokking, producing both the large lag
and the $\approx 3\times$ transient). The single LayerNorm ablation B-vs-C moves frac-pre from $0.87$ to
$0.25$: within this harness it is LayerNorm, not ``being a transformer,'' that defers compression past
grokking. Equivalently, the grok-to-compression lag is smallest for the canonical transformer and large
with LayerNorm or on the MLP --- but present, not zero, in every arm (the canonical transformer still
settles within $\approx 0.63\,T_{\mathrm{grok}}$ of grokking, Sec.~\ref{sec:transformer}). We read the lag
qualitatively here, since its precise coefficient is seed-sensitive when a high frac-pre leaves only a
narrow band to the floor, and note that its direction agrees with the better-powered transformer estimate
of Section~\ref{sec:transformer} ($\mathrm{lag}/T_{\mathrm{grok}}\approx 0.63$). The transient and the
lag are thus two views of one quantity --- how much of the representation has compressed at the moment
one reads it; the normalization scheme modulates that quantity. Consistent with this, the post-grok
coupling between the global weight norm and the embedding rank is strongly negative only under LayerNorm
(Spearman $-0.80$) and positive for the canonical model ($+0.70$): the LayerNorm arm compresses as its
norm grows after grokking, whereas the canonical arm has already compressed.

\paragraph{A note on power.} This harness changes one variable cleanly but is not high-powered: it uses
few seeds and a $6\times10^4$-step budget, and the most dramatic magnitude (the $\approx 3.2\times$
LayerNorm transient) comes from it, whereas the well-powered transformer sweep of
Section~\ref{sec:transformer}, trained to true convergence, gives a more modest $1.3$--$1.5\times$
embedding transient. We therefore state the mechanism --- normalization modulates frac-pre, and frac-pre
orders the size of the lag --- as the contribution, and scope the magnitudes: the direction is consistent
across the well-powered sweep and this one-variable harness, but the precise coefficients
($\mathrm{lag}/T_{\mathrm{grok}}$, the transient multiple) are not pinned down by the present data.

\begin{figure}[H]
\centering
\includegraphics[width=\textwidth]{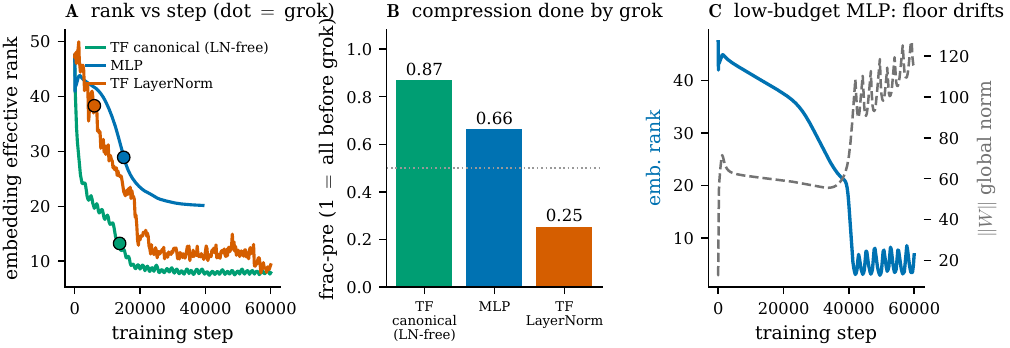}
\caption{One harness, three architectures, one variable at a time. Embedding squared-normalized effective
rank under free weight decay, $p=59$. \textbf{(A)} Representative rank trajectories with the grok step
marked (dot): the canonical transformer (green) has compressed almost entirely by grokking, the
LayerNorm transformer (red) compresses after grokking, and the MLP (blue) is intermediate. \textbf{(B)}
Fraction of the embedding-rank compression completed by the grok step (frac-pre): $0.87$ canonical,
$0.66$ MLP, $0.25$ LayerNorm --- the single quantity that orders the arms, and equivalently sets whether
the lag is near-zero or large. \textbf{(C)} A low-budget MLP cell ($\mathrm{wd}=0.5$): after grokking the
global norm doubles and the rank undergoes a second, late collapse, so the final-tenth ``floor'' is
itself transient.}
\label{fig:harness}
\end{figure}

\paragraph{A second non-stationarity: the floor itself need not be stable.} The low-budget MLP cells
($\mathrm{wd}=0.5$) expose a separate hazard that sharpens the paper's thesis. There the embedding groks
at rank $\approx 22$, holds a plateau, and then, well after grokking, undergoes a second transition in
which the global weight norm roughly doubles and the rank collapses to $\approx 4$ (Fig.~\ref{fig:harness}C).
The converged floor, computed over the final tenth of training, is read inside this second regime, so a
naive floor is itself transient for these cells. The same failure the audit is built to catch on the
at-grok reading can therefore recur on the denominator: a careful audit must verify that the floor has
plateaued, not merely that training has run long. The audit's floor-plateau check exists for exactly this
case.

\paragraph{What this revises.} This harness refines the transformer lag of Section~\ref{sec:transformer}
rather than overturning it: that section's un-normalized model is arm B here, and its low-power lag
estimate agrees in direction with the frac-pre ordering measured here --- a canonical transformer
compresses at or near grokking, not long after it. The ``large lag comparable to $T_{\mathrm{grok}}$''
that the MLP shows (Sec.~\ref{sec:lag}) and that a LayerNorm transformer shows is therefore
normalization-mediated, not a property of grokking as such. A concurrent empirical study reinforces this
picture. \citet{manir2026systematic} find, in a controlled study on modular addition, that the apparent
Transformer--MLP grokking gap largely dissolves (a $1.11\times$ delay) under matched hyperparameters, and
they attribute previously reported architecture differences to optimization and regularization
confounds. Our one-variable ablation localizes one such factor --- normalization --- as a specific,
controllable variable inside that picture. The audit's job is unchanged: report the separation that
exists and name the variable, normalization, that the data identify, without universalizing a coefficient
that the present data do not pin down.

\begin{figure}[H]
\centering
\includegraphics[width=\textwidth]{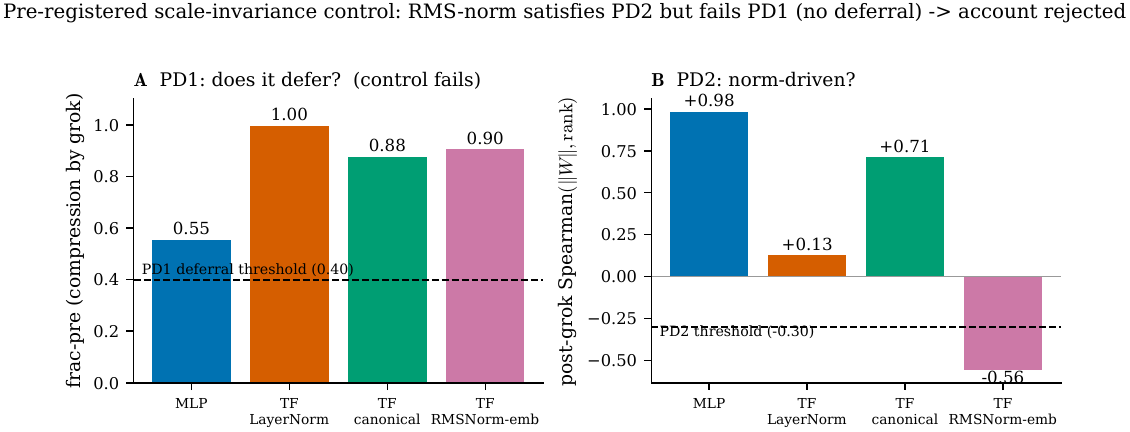}
\caption{The pre-registered scale-invariance control (thresholds frozen before the run). Adding
RMS-normalization on the embedding input only (\texttt{tf\_rms}: per-token scale invariance, no
mean-centering, affine gain, or per-sublayer norm) does not reproduce LayerNorm's deferral. \textbf{A}
PD1 (deferral): frac-pre for \texttt{tf\_rms} is $0.90$, above the $0.40$ threshold and indistinguishable
from the canonical transformer ($0.88$) --- compression is largely complete by grokking, not deferred
like LayerNorm. \textbf{B} PD2 (norm-driven): \texttt{tf\_rms} does show a negative post-grok coupling
between the global weight norm and the embedding rank ($-0.56$, below the $-0.30$ threshold). Acceptance
requires PD1 \emph{and} PD2; PD1 fails, so the scale-invariance account is rejected.}
\label{fig:scaleinv}
\end{figure}

\paragraph{The scale-invariance account, pre-registered and then rejected.} Why LayerNorm defers
compression and the canonical transformer does not is a mechanism question, and a natural hypothesis is
scale invariance: LayerNorm makes the network invariant to the scale of its inputs, so it could reach
correct predictions while the embedding is still high-rank, deferring compression to a later,
norm-driven phase. Because this is exactly the kind of mechanism one can talk oneself into after the
fact, we fixed the falsifier in advance (\texttt{specs/PREREGISTRATION\_scale\_inv.json}, thresholds
frozen before the run): a control arm \texttt{tf\_rms} that adds RMS-normalization on the embedding input
only --- per-token scale invariance, with no mean-centering, affine gain, or per-sublayer norm --- run
under the same harness, with the account accepted iff both PD1 (deferral: $\mathrm{frac\text{-}pre} <
0.40$) \emph{and} PD2 (norm-driven: post-grok $\mathrm{Spearman}(\lVert W\rVert,\mathrm{rank}) < -0.30$)
hold for \texttt{tf\_rms}. Running the control \emph{rejects} the account (Fig.~\ref{fig:scaleinv}):
\texttt{tf\_rms} satisfies PD2 (post-grok Spearman $=-0.56$) but decisively fails PD1, with
$\mathrm{frac\text{-}pre}=0.90$ --- compressing \emph{before} grokking, like the canonical transformer
($0.88$), not deferring like LayerNorm. Per-token input-scale invariance is therefore not what defers
compression; whatever LayerNorm does to enlarge the lag is not captured by it. This has a concurrent
theoretical counterpart. \citet{li2026lowrank} observes that under exact scale invariance, Frobenius
weight decay acts purely along the radial direction, so once task gradients vanish it cannot reshape a
normalized layer's singular-value spectrum. Scale invariance thus changes \emph{how} decay compresses
rank, not \emph{whether} it does. Our control provides the empirical complement: it isolates per-token
input-scale invariance on the embedding and finds this is not the operative channel for the observed
deferral. We report this as a pre-registered negative result: the audit's discipline applies to our own
most natural mechanistic explanation, and falsifies it. That a negative post-grok norm--rank coupling
(PD2) can hold \emph{without} deferral (PD1) is itself a caution --- the discriminating prediction is the
deferral one, and a norm-driven signature alone would have been a misleading confirmation.

\section{A transformer dose-response, and a depth-law generality test}
\label{sec:transformer}
The MLP results above raise two questions that the same intervention can answer on a second
architecture: does the at-grok transient, the paper's central claim, generalize, and does the norm-budget
depth law (the secondary, MLP-specific observation of Appendix~\ref{app:depthlaw}) generalize? We
therefore run a well-powered transformer sweep to true convergence, a one-layer un-normalized attention
model on modular addition ($p=59$), under both the clamp protocol (seven budgets $\rho\in[0.9,1.4]$) and
a free weight-decay protocol (six decays), three seeds each, with dense post-grok logging of five
spectral measures at four weight loci and the activation representation. The two questions settle in
opposite directions.

\begin{figure}[H]
\centering
\includegraphics[width=\textwidth]{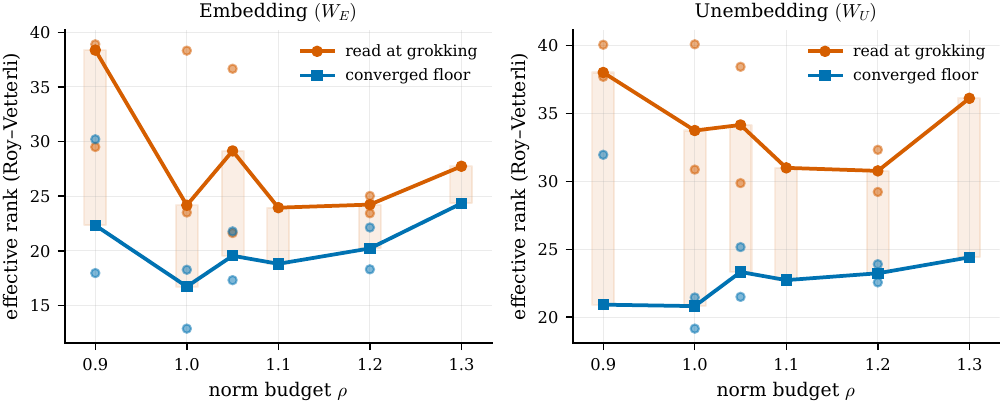}
\caption{The at-grok transient generalizes to the transformer. Effective rank (Roy--Vetterli) read at
grokking (red) vs.\ the converged floor (blue) across the clamp norm budget $\rho$, for the embedding
$W_E$ (left) and unembedding $W_U$ (right); points are generalizing, non-censored seeds and lines are
per-budget medians, with the transient shaded. The at-grok value sits above the converged floor at almost
every budget on both loci (median $1.34\times$ on $E$, $1.48\times$ on $U$), reproducing the modular-MLP
transient on a different architecture at a smaller magnitude.}
\label{fig:tf}
\end{figure}

\paragraph{The central claim generalizes: the transient is not specific to the MLP.} On the clamp sweep,
the embedding effective rank read at grokking sits above its converged floor in $92\%$ of generalizing,
non-censored cells (median $1.34\times$); the unembedding overstates by $1.48\times$ in every such cell,
and the free-decay sweep shows the same ($1.36\times$ and $1.38\times$; Fig.~\ref{fig:tf}). The
overstatement is metric-agnostic, appearing in Roy--Vetterli and variance-normalized effective rank,
participation ratio, and stable rank alike, and present at every locus we log: it is largest on the
MLP-block output weights ($W_{\mathrm{out}}$, up to $3.6\times$ at the smallest budget), confirming that
the hazard is a property of representation compression rather than of one matrix or one measure. As on
the MLP, the size of the transient is set by the budget (largest at small $\rho$, shrinking toward the
high-budget cells). The magnitude is more modest than the $3$--$5\times$ overstatement of the modular
MLP: on the transformer the embedding transient is $\approx 1.3$--$1.5\times$. We report it for what it
is, a smaller but robust hazard that now recurs on a second architecture, and do not inflate the headline
magnitude.

\paragraph{The lag recurs.} The post-grok compression lag also transfers: across the clean transformer
cells the median lag is $\approx 2.7\times10^4$ steps ($\mathrm{lag}/T_{\mathrm{grok}}\approx 0.63$), and
as on the MLP the compression time is not cleanly ordered by the budget
($\rho_S(\rho,\mathrm{lag})=+0.37,\, p=0.47,\, n=6$ cells --- a low-power estimate, since several clean
cells rest on a single generalizing seed, so we read it as a non-ordering rather than a precise
coefficient). The large-but-not-budget-ordered separation of generalization onset from representation
compression is thus a consistent two-architecture finding. This lag is computed by the same per-seed
clock as the MLP audit, restricted to the generalizing, non-censored clamp cells.

\paragraph{The depth law does not generalize (a negative result; full detail in Appendix~\ref{app:depthlaw}).}
The norm-budget depth law is a different story, and we report the negative result plainly. On the
transformer the converged floor does not fall with the budget at any weight or activation locus we log:
on the embedding it is essentially unordered ($\rho_S(\rho,\mathrm{floor})=+0.14$, $p=0.66$), on the
unembedding weakly positive ($+0.42$, n.s.), and on the activation representation significantly positive
($+0.82$, $p<0.01$). Where the modular MLP shows a strict monotone fall ($-1.0$), the transformer floor
is flat or, on the activation locus, ordered in the opposite direction. Combined with the protocol
reversal on the MLP (free weight decay orders the floor oppositely to the clamp; Appendix~\ref{app:depthlaw}),
this establishes that the depth law is specific to the modular-arithmetic MLP and does not survive a
change of architecture. The standard that motivates this paper applies to its own secondary result: a
clean dose-response on one architecture is not a law until it is shown on another, and here it is not.

\paragraph{Generalization is seed-fragile here, and we flag it.} The un-normalized transformer groks
erratically across seeds at this configuration: only 16 of 21 clamp cells and 14 of 18 free-decay cells
reach $0.90$ test accuracy, with grok times for the same budget ranging from $\sim 10^3$ to $>10^5$ steps
and one seed failing to grok at all. We therefore restrict every transformer statistic above to cells
that fully generalize and whose floor has plateaued, and we report the transformer as a generalization
check on the transient (which passes) and a generality test of the depth law (which fails), not as a
clean second dose-response.

\section{Why a tested audit, and not just a metric}
The case for this paper is not a new metric, since effective rank is borrowed, but the claim that
measurement reliability is itself a result in a literature that increasingly reads structure off single
transition-time snapshots. We make that case concretely. The audit is built around explicit failure
modes rather than a happy path. It (i) flags censoring when cells do not reach the compression threshold
within budget; (ii) excludes boundary/incomplete-grok cells before any ordering statistic is computed;
(iii) emits \textsc{low-power / descriptive only} when fewer than two non-boundary cells remain; (iv)
verifies that the compression floor has itself plateaued before using it as a reference; and (v) returns
\textsc{large lag, ordering undetermined} --- rather than asserting a shared clock --- whenever the order
statistic across budgets is undefined (too few cells or tied values). Each of these is a verdict the tool
will return against a tidy conclusion when the data do not support one; in our own modular-multiplication
data the weak ($+0.40$) ordering (on the earlier, smaller grid) is exactly the regime where this
restraint matters.

\paragraph{Why this matters beyond effective rank, and the three axes it lives on.} The hazard is not a
property of effective rank in particular but of reading any structural summary at the transition. The
same transition-time checkpoint at which one reads embedding rank is also where a Fourier decomposition,
a persistent-homology diagram, or an intrinsic-dimension estimate of the representation would be read ---
and each of those is a quantity still in motion at $T_{\mathrm{grok}}$; indeed a Fourier
circuit-synchronization measure and $H_1$ persistence move in the opposite direction to rank across the
transition \citep{sivasankar2026circuit,tang2026topological}. A mechanistic-interpretability pipeline that
characterizes the grokked circuit from a transition-time snapshot therefore inherits the transient on
whatever metric it uses. More generally, three choices determine what such a reading reports, and each
varies independently across the literature: which quantity is measured (rank falls, while circuit
synchronization and topological persistence rise), at which locus (the embedding, the first layer, and
their product compress on different schedules \citep{alquabeh2025embedding}), and when it is read (the
onset-to-compression lag we date). Our audit fixes the last axis --- it decides whether a reading has
settled in time --- and its metric-agnostic input schema is built to range over the first two.

\paragraph{The audit caught its own author.} The strongest evidence that these checks are load-bearing is
that they fired on us. During development a reporting branch that added genuinely useful low-power and
no-grok diagnostics was forked from before two ordering fixes, and it re-introduced a verdict that
asserted ``one clock'' while the underlying rank correlation was undefined ($\mathrm{NaN}$ Spearman). The
nine-case adversarial suite caught this, not because a verdict string was missing, but because the suite
requires a clock-type verdict to be backed by a finite order statistic. A test that checked only for the
presence of the words ``one clock'' would have passed the broken analyzer; its own bundled test did. This
is one way a representation study can fool itself, with a measurement that looks like a finding but rests
on an undefined or transient quantity. The same check fired a second time at the data level: an apparent
single-clock ordering on a small budget grid dissolved once the grid was enlarged, and we report the
corrected reading rather than the original (Sec.~\ref{sec:lag}).

\paragraph{Adversarial test suite.} The suite ships nine pre-registered cases: clean one-clock,
all-censored, high-floor boundary, true two-clock, single-valid-cell, non-monotone (rank rebounds after
falling), compression-before-grok, tied orderings, and heavy noise. Each case fixes the expected verdict
and the reason before running; a case that passes only by avoiding a crash is treated as a failure. The
suite, the analyzer, an adapter for externally-logged runs, sample data, and the figure generator are
released (Appendix~\ref{app:toolkit}) so that the audit can be run on other studies' trajectories.

\section{Takeaways and future development}
\paragraph{Takeaways.} The paper's contribution is not a new representation metric but a shift in how an
existing one is read: that a low-rank reading taken at the grokking transition can be a transient that
does not reflect the converged circuit, that the onset-to-compression lag is itself a measurable quantity
rather than a control parameter, and that this is packaged as a tested audit rather than asserted in
prose. Concretely: (1) On modular arithmetic, embedding effective rank at grokking is a transient that
long training drives to a low-rank floor; it shows up in participation ratio and stable rank as well, and
recurs on the transformer (Sec.~\ref{sec:transformer}), so a transition-time snapshot can misrepresent
the converged circuit. (2) Rank compression lags the accuracy transition rather than coinciding with it
(contra a common reading) by an amount of order $T_{\mathrm{grok}}$ on the MLP; the separation is the
robust, replicated effect, and its magnitude is normalization-dependent (Sec.~\ref{sec:modulates}). (3)
The variable that modulates the size of the lag is normalization: a post-grok lag is present on every
architecture, and a one-variable LayerNorm ablation moves frac-pre from $0.87$ to $0.25$, enlarging it.
The mechanism is narrowed but not closed: a pre-registered control rejects per-token scale invariance as
the cause ($\mathrm{frac\text{-}pre}=0.90$ for the RMS-norm arm, like the canonical transformer, not
deferred; Sec.~\ref{sec:modulates}). (4) The norm budget orders $T_{\mathrm{grok}}$ but not
$T_{\mathrm{compress}}$. A secondary observation, held to a higher bar and then broken ourselves: on the
MLP the budget sets the depth of compression (converged effective-rank floor Spearman $-1.0$), but this
depth law is MLP-specific ($+0.14$ vs.\ $-1.0$ on the transformer; Sec.~\ref{sec:transformer}) and
protocol-specific (Appendix~\ref{app:depthlaw}), so we report it as a negative result on generality, not
a norm law. (5) Measurement reliability is demonstrable, not assumed: the adversarial suite and the
enlarged grid each corrected a reading of our own; and the hazard recurs on the denominator (the floor
can be transient too), which the floor-plateau check guards.

\paragraph{Future development of the toolkit.} Several extensions are natural and low-cost.
\emph{Same-run multi-clock measurement}: add persistent-homology, LID, and Fourier-Gini clocks to the
analyzer so the rank, topology, and feature clocks are placed on one timeline from a single run, the
experiment that would turn our rank-vs-topology hypothesis into a result. \emph{Checkpoint-density
guidance}: a planned ablation that subsamples post-grok checkpoints to report how many are needed for a
reliable $T_{\mathrm{compress}}$, turning the transformer censoring we observed into a concrete logging
recommendation. \emph{Metric-agnostic clocks}: the at-grok transient is already metric-agnostic,
appearing in participation ratio and stable rank as well as effective rank (Sec.~\ref{sec:atgrok}), and
since the clock definition needs only a quantity that falls as the representation compresses, running the
full lag measurement under each of these measures is a direct extension; the analyzer's input schema
accepts any such quantity. \emph{Locus selection on transformers}: promote the two-point E-vs-U locus
check into a full per-locus clock, since the right compression locus on a transformer may be the
unembedding rather than the embedding. \emph{A metric-timing atlas}: because different representational
quantities cross grokking in opposite directions --- a Fourier circuit-synchronization measure leads it
\citep{sivasankar2026circuit} while embedding effective rank lags it (this work) --- placing several such
clocks on one timeline from a single run would turn ``compression relative to grokking'' from a single
number into a per-metric map, and is exactly what the analyzer's metric-agnostic input schema is built to
support. \emph{Decomposing the normalization mechanism}: having pre-registered and rejected per-token
input-scale invariance as the cause of the LayerNorm-mediated lag, the natural next controls isolate the
remaining LayerNorm components one at a time --- mean-centering, the affine gain, and per-sublayer
placement --- against the theoretical prediction that scale invariance reshapes how weight decay acts on
the spectrum \citep{li2026lowrank}. \emph{Connection to phase-transition accounts}: grokking has been
framed as a first-order transition with metastable trapping and hysteretic escape
\citep{ersoy2026noise}; whether the post-grok compression event we date corresponds to a distinct, later
barrier crossing than the accuracy transition itself is a question the same two-clock machinery could
address. \emph{Forward/backward coupling}: pairing the audit with a forward detector
\citep{wang2026spectral,xu2026earlywarning,golwala2026ildr} would let a pipeline both predict a
transition and certify whether a measurement taken at it has converged. \emph{Does any budget-to-floor
relation survive scale?} The monotone budget-to-floor relation is established on a small modular MLP, and
the first generality test, a second architecture, already comes back negative
(Sec.~\ref{sec:transformer}); whether any budget-to-floor relation reappears on the larger, non-modular
systems where rank minimization has been shown to operate (VGG, U-Net, LSTM, language-model transformers;
\citep{yunis2024spectral}) and across the model scales over which weight decay structures grokking
regimes \citep{verma2026weightdecay} is open. The audit's third-party adapter is built to run that test on
others' checkpoints without re-training.

\section{Limitations}
The full compression-lag dose-response is established on the MLP (two tasks, seven and nine norm
budgets); on the enlarged grid the compression time is not ordered by the norm budget
($\rho_S(\rho,T_{\mathrm{compress}})=+0.18,+0.29$), so we report the large lag (robust and replicated
across the architectures we tested) and the monotone floor law (robust on the MLP, but
architecture-specific) as the two MLP findings, and explicitly do not claim a single shared clock.

\paragraph{What we do and do not attribute to prior work.} We do not exhibit a published study that read
at-grok representation structure as converged and thereby reached a reversed conclusion, and we
attribute no such error to specific prior work: the careful complexity, topology, and
``construct-then-compress'' accounts track the full trajectory. Our claim is the weaker but verifiable
one that the single-snapshot version of a common practice --- reading a representation-complexity metric
at the transition and comparing it across a control parameter
\citep{tang2026topological,yunis2024spectral,brown2022intrinsic} --- is unguarded against the transient we
document, and that the transient is real rather than an artifact of our metric or setup: it has been
reported independently, as a transient rise in the intrinsic dimension of activations at the onset of
grokking, on a different task \citep{brown2022intrinsic}. The strongest concrete timing target we name is
the reading ``low-rank compression coincides with grokking'' \citep{yunis2024spectral}, which our
long-train lag corrects.

\paragraph{The audit can have the disease it diagnoses, and we guard it explicitly.} The reference floor
against which the at-grok value is judged can itself be a transient, as the $\mathrm{wd}=0.5$ MLP cells
show (a second, late collapse after a plateau; Sec.~\ref{sec:modulates}). This is the same hazard on the
denominator, and it means the tool does not solve convergence: it detects one species of non-convergence
(a metric still falling at the read step) and flags censoring, and it requires a floor-plateau check that
we make part of the verdict. We do not claim a general certificate that a representation has converged;
we claim a tested procedure that catches the specific, common way a transition-time reading misleads, on
both the numerator and the denominator.

\paragraph{The normalization-mediated lag is a mechanism claim at modest power.} The one-variable harness
(Sec.~\ref{sec:modulates}) isolates LayerNorm cleanly but uses few seeds and a $6\times10^4$-step budget;
its most dramatic magnitude (the $\approx 3.2\times$ LayerNorm transient) is the least well-powered
number in the paper. We state the direction --- normalization defers compression and thereby enlarges
the lag --- as the finding, corroborated by the well-powered transformer sweep's agreeing direction
($\mathrm{lag}/T_{\mathrm{grok}}\approx 0.63$; Sec.~\ref{sec:transformer}), and we do not present the
precise coefficients as pinned down.

\paragraph{The depth law is intervention-specific.} The depth law itself is established only on a small
modular MLP at one width and is, at most, a modular-MLP-specific echo of a qualitative observation of
\citet{yunis2024spectral} rather than a general quantification of it; it does not generalize across
architecture (a transformer clamp sweep trained to convergence shows no monotone floor law;
Sec.~\ref{sec:transformer}) and its direction is protocol-specific (a free weight-decay sweep at $p=59$,
three tasks including parity, shows the converged floor rising monotonically with decay strength,
$\rho_S(\mathrm{wd},\mathrm{floor})=+1.0$, the opposite sign to the clamp budget; Appendix~\ref{app:depthlaw}).
That the two protocols disagree on direction is not in itself a contradiction, because they hold
different quantities fixed: the clamp imposes a global norm throughout training, whereas free weight
decay lets the norm equilibrate against the loss. We report the depth law as a property of the norm-clamp
intervention rather than a protocol-independent norm law; the transient itself, by contrast, replicates
across both protocols and all three tasks (Sec.~\ref{sec:atgrok}).

\paragraph{What we do not measure.} We do not measure persistent homology or intrinsic dimension on the
same runs, so the rank-vs-topology timing contrast with \citet{tang2026topological} is a hypothesis, not
a result, and the cross-paper lag ratio is suggestive only (different setups). The effective-rank metric
and the norm-clamp protocol are borrowed, not introduced; the contribution is the audit and the procedure
around it. We make no mechanistic claim about \emph{why} compression lags grokking beyond identifying
normalization as the controlling variable; we provide a reliable way to measure that the lag exists, what
sets its size, and to avoid mistaking the transient for the endpoint. Finally, since both generalization
onset and the subsequent norm contraction are driven by weight decay, any account in which the delay is
set by the weight norm --- the norm picture of \citet{liu2023omnigrok}, and our own norm-separation delay
law \citep{truong2026normsep} --- predicts the compression lag to scale with the same norm budget that
sets $T_{\mathrm{grok}}$; testing whether the lag obeys the delay law's scaling is a concrete next step
the clock is built to support.

\section{Conclusion}
Reading representation structure off a snapshot at the grokking transition is now common, and it can
mislead, because at-grok structure can be a transient that the converged network discards --- and, we
show, the reference floor can be transient too. We provide a measurement-validity audit that separates
generalization onset from representation compression, handles censoring and boundary cells, verifies
that the floor has plateaued, refuses to claim an ordering it cannot support, and is checked by an
adversarial test suite that fails on false confidence. On modular arithmetic the audit shows that
generalization precedes low-rank compression by a large but non-independent lag, and that the at-grok
rank peak is transient. A one-variable architectural ablation then identifies what controls the lag:
normalization, not grokking as such, sets how much of the representation has compressed by the time one
reads it. A well-powered transformer sweep carries the transient to a second architecture and shows the
norm-budget depth law to be MLP-specific, a reported negative result. We release the analyzer, the
adapter for externally-logged runs, the test suite, sample data, and the figure-generation script, so
that other grokking studies can check whether their transition-time measurements describe the converged
circuit or a passing state.

\paragraph{Code and data.} \url{https://github.com/ClevixLab/grokking-compression-clock}

\clearpage
\appendix

\section{The norm-budget depth law (MLP-specific, self-falsified)}
\label{app:depthlaw}
This appendix collects the secondary observation that the norm budget sets the depth of compression on
the modular MLP, and the two generality tests it fails. We present it as a scoped negative result on
generality, held to a higher bar than the central transient and lag claims of the main text, and broken
by our own pre-registered tests.

\begin{figure}[H]
\centering
\includegraphics[width=\textwidth]{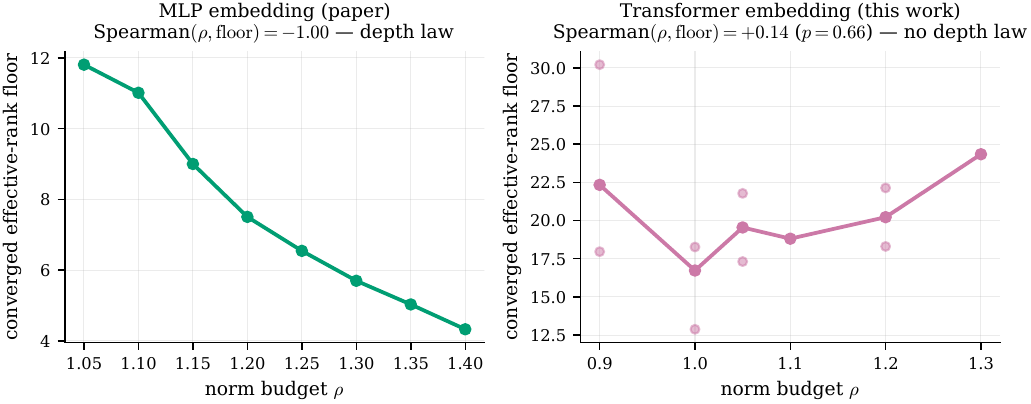}
\caption{The norm-budget depth law is MLP-specific (negative result). Converged effective-rank floor
vs.\ the clamp norm budget $\rho$. Left, the modular-addition MLP (paper data): the floor falls
monotonically, Spearman $-1.0$ --- the depth law. Right, the transformer embedding (non-censored
generalizing seeds): the floor does not fall, Spearman $+0.14$ ($p=0.66$). The relationship that looked
like a law on the MLP does not replicate on a second architecture.}
\label{fig:depthlaw}
\end{figure}

\paragraph{What the norm budget controls on the MLP: the depth of compression.} On the MLP, across the
grid the norm budget sets, almost deterministically, the depth of compression rather than its timing:
the converged effective-rank floor decreases monotonically with $\rho$ ($\rho_S(\rho,\mathrm{floor})=-1.0$
at both tasks, bootstrap $95\%$ CI $[-1.0,-1.0]$ on addition and $[-1.0,-0.93]$ on multiplication), and
the rank read at grokking is itself lower at higher budgets ($\rho_S\approx -0.97$). This explains the
weak ordering of $T_{\mathrm{compress}}$: a higher-budget cell groks later but starts post-grok
compression from an already-lower rank with less distance to fall. We do not over-read this depth law.
It is an MLP-specific regularity that does not survive a change of architecture, as the transformer sweep
shows (Sec.~\ref{sec:transformer}, Fig.~\ref{fig:depthlaw}), and we report it as a scoped observation
rather than a norm law.

\paragraph{Generality test 1 (architecture): fails on the transformer.} On the transformer the converged
floor does not fall with the budget at any weight or activation locus we log: on the embedding it is
essentially unordered ($\rho_S(\rho,\mathrm{floor})=+0.14$, $p=0.66$, non-censored generalizing cells), on
the unembedding weakly positive ($+0.42$, n.s.), and on the activation representation significantly
positive ($+0.82$, $p<0.01$). Where the modular MLP shows a strict monotone fall ($-1.0$; Fig.~\ref{fig:depthlaw}),
the transformer floor is flat or, on the activation locus, ordered in the opposite direction. The
negative result therefore does not hinge on a single non-significant null on one matrix: it is a
consistent failure of the monotone fall across three loci, one of which reverses the sign outright. We
checked that this is not a censoring artifact: high-budget cells grok late and were excluded if their
floor was still falling at the step cap, and the result holds on the cells that remain.

\paragraph{Generality test 2 (protocol): the sign reverses under free weight decay.} The depth law's
direction is protocol-specific. A free weight-decay sweep at $p=59$ (three tasks, including parity) shows
the converged floor rising monotonically with the decay strength ($\rho_S(\mathrm{wd},\mathrm{floor})=+1.0$
on the modular tasks), the opposite sign to the clamp budget, and the two protocols do not collapse onto
a common floor-versus-converged-norm curve ($\rho_S(\mathrm{norm},\mathrm{floor})=+0.4$ under free decay
vs.\ $-1.0$ under the clamp). That the two protocols disagree on direction is not in itself a
contradiction, because they hold different quantities fixed: the clamp imposes a global norm throughout
training, whereas free weight decay lets the norm equilibrate against the loss, so the former measures
how a representation arranges a prescribed budget and the latter the budget the dynamics select. Whether
the clamp depth law is a property of the generalizing solution or of the imposed-norm intervention is
settled in the negative by generality test 1: the second-architecture clamp run trained to convergence
shows the converged floor does not order with the budget on the transformer ($\rho_S=+0.14$ on the
embedding, versus $-1.0$ on the MLP), so the depth law is MLP-specific.

\section{Sensitivity to the audit's free parameters}
\label{app:sensitivity}
The audit exposes five thresholds: the compression fraction $\varepsilon$, the floor-averaging window,
the grokking accuracy threshold, the minimum at-grok drop, and the boundary drop-threshold.
Table~\ref{tab:sensitivity} varies each one at a time, holding the rest at their defaults. Three findings
are invariant across every setting: the converged floor law is Spearman $-1.0$ on both tasks; the
boundary gate excludes exactly the $\rho=1.00$ cell on addition and no cell on multiplication; and the
large post-grok lag is always retained while the $T_{\mathrm{grok}}/T_{\mathrm{compress}}$ ordering never
flips to opposite-signed --- so the verdict never collapses to a coincident (zero-lag) shared clock and
never becomes two oppositely-ordered clocks. In 28 of the 30 settings the verdict is \textsc{partially
separated + large lag}; in 2 ($\varepsilon=0.20$ on multiplication and grok thr $=0.99$ on addition) the
same-signed ordering is strong enough that the analyzer labels it \textsc{one clock + large lag}, i.e.\ a
single ordering with the lag still present, not a coincident clock --- the verdict family is stable even
where the label crosses that threshold. The one quantity that moves is the lag coefficient:
$\mathrm{lag}/T_{\mathrm{grok}}$ ranges over $[0.65,1.46]$ as $\varepsilon$ varies from $0.20$ to $0.05$,
because $\varepsilon$ sets how close to the floor counts as ``compressed.'' The lag magnitude stays of
order $T_{\mathrm{grok}}$ ($\ge 10^4$ steps) throughout; only its calibration moves. We report
$\varepsilon=0.10$ (where $\mathrm{lag}/T_{\mathrm{grok}}\approx 1.0$) as the default and note that this
dependence is monotone and expected, not a sign of fragility.\footnote{The $\mathrm{lag}/T_{\mathrm{grok}}$
column of this sensitivity table is computed as the ratio of medians (median gap over median
$T_{\mathrm{grok}}$ across cells), giving $1.03$ for both tasks at $\varepsilon=0.10$; Table~\ref{tab:clocks}
instead reports the canonical analyzer's median of the per-cell ratios ($1.00$ on addition, $1.04$ on
multiplication). The two aggregations agree to within $0.04$, and the tolerance-independent claim (lag
$\ge 10^4$ steps, of order $T_{\mathrm{grok}}$) does not depend on either choice.}

\begin{table}[H]
\centering
\scriptsize
\setlength{\tabcolsep}{3pt}
\caption{Sensitivity of the audit to its free parameters (full grid; each parameter varied one at a
time, the rest held at the defaults $\varepsilon=0.10$, floor frac $0.10$, grok thr $0.90$, min drop
$1.0$, gate thr $0.25$). The converged-floor law $\rho_S(\mathrm{floor})=-1.0$ and the boundary gate
(addition excludes $\rho=1.00$; multiplication excludes none) are invariant across all settings; only the
lag coefficient $\mathrm{lag}/T_{\mathrm{grok}}$ moves, tracking $\varepsilon$ as expected.}
\label{tab:sensitivity}
\begin{tabular}{llccccc}
\toprule
parameter & task & lag & $\mathrm{lag}/T_{\mathrm{grok}}$ & $\rho_S(T_{\mathrm{grok}})$ &
$\rho_S(T_{\mathrm{compress}})$ & $\rho_S(\mathrm{floor})$ \\
\midrule
$\varepsilon=0.05$ & add & 24750 & 1.36 & +0.85 & +0.17 & $-1.00$ \\
$\varepsilon=0.05$ & mult & 25500 & 1.46 & +0.90 & +0.39 & $-1.00$ \\
$\varepsilon=0.10$ & add & 18000 & 1.03 & +0.85 & +0.29 & $-1.00$ \\
$\varepsilon=0.10$ & mult & 17000 & 1.03 & +0.90 & +0.18 & $-1.00$ \\
$\varepsilon=0.20$ & add & 11500 & 0.65 & +0.85 & +0.40 & $-1.00$ \\
$\varepsilon=0.20$ & mult & 10500 & 0.67 & +0.90 & +0.20 & $-1.00$ \\
floor frac $=0.05$ & add & 18000 & 1.03 & +0.85 & +0.29 & $-1.00$ \\
floor frac $=0.05$ & mult & 17000 & 1.03 & +0.90 & +0.21 & $-1.00$ \\
floor frac $=0.20$ & add & 18000 & 1.03 & +0.85 & +0.29 & $-1.00$ \\
floor frac $=0.20$ & mult & 17000 & 1.03 & +0.90 & +0.18 & $-1.00$ \\
grok thr $=0.95$ & add & 18750 & 0.97 & +0.91 & +0.31 & $-1.00$ \\
grok thr $=0.95$ & mult & 17500 & 0.97 & +0.92 & +0.32 & $-1.00$ \\
grok thr $=0.99$ & add & 20000 & 0.93 & +0.91 & +0.30 & $-1.00$ \\
grok thr $=0.99$ & mult & 19500 & 0.97 & +0.98 & +0.54 & $-1.00$ \\
min drop $=0.5$ / $2.0$ & add & 18000 & 1.03 & +0.85 & +0.29 & $-1.00$ \\
min drop $=0.5$ / $2.0$ & mult & 17000 & 1.03 & +0.90 & +0.18 & $-1.00$ \\
gate thr $=0.20$ / $0.30$ & add & 18000 & 1.03 & +0.85 & +0.29 & $-1.00$ \\
gate thr $=0.20$ / $0.30$ & mult & 17000 & 1.03 & +0.90 & +0.18 & $-1.00$ \\
\bottomrule
\end{tabular}
\end{table}

\section{Toolkit catalog}
\label{app:toolkit}
The release is a small, self-contained audit toolkit. The design separates the clock logic (one
analyzer, versioned and tested) from the input adaptation (an adapter for externally-logged runs) and
the evidence (a test suite, a figure generator, and sample data), so the same audit logic runs unchanged
across run formats. Each component is listed below with its role.

\begin{description}
\item[\ttfamily analyze\_compression\_clock\_v1\_5.py] \emph{(analyzer).} Canonical clock. Computes
$T_{\mathrm{grok}}$, $T_{\mathrm{compress}}$, lag, floor, and censoring per norm budget; applies the
boundary gate; emits the verdict (one-clock / two-clock / partially-separated /
large-lag-ordering-undetermined / low-power / censoring-inconclusive). Writes a CSV, a per-seed CSV, and
a provenance JSON.
\item[\ttfamily analyze\_compression\_clock\_generic\_v1.py] \emph{(third-party adapter).} Runs the same
audit on runs that are not ours: it takes either a directory of weight checkpoints (computing the
effective rank of a named matrix at each logged step) or a logged metrics table, normalizes it to the
analyzer's schema, and reports the verdict --- so others can audit their own grokking runs without
re-training.
\item[\ttfamily transient\_replication.py, metric\_agnostic\_transient.py] \emph{(Sec.~\ref{sec:atgrok}).}
Reproduce the free-weight-decay transient on modular addition, multiplication, and parity, and check it
across participation ratio and stable rank in addition to effective rank.
\item[\ttfamily corrected\_reading.py] \emph{(Sec.~\ref{sec:lag}).} Recomputes the $T_{\mathrm{compress}}$
ordering statistics on both the original small budget grid and the enlarged grid, so the corrected
reading is auditable rather than asserted.
\item[\ttfamily sensitivity\_floor\_ci.py] \emph{(Appendix~\ref{app:sensitivity}).} Sweeps the audit's five
free thresholds one at a time and bootstraps the floor-law confidence interval.
\item[\ttfamily arch\_generality\_audit.py, h1\_frac\_pre\_bootstrap.py] \emph{(Sec.~\ref{sec:modulates}).}
Run the one-variable MLP / canonical-transformer / LayerNorm-transformer harness and compute frac-pre with
bootstrap confidence intervals.
\item[\ttfamily analyze\_scale\_invariance.py] \emph{(Sec.~\ref{sec:modulates}).} Runs the pre-registered
PD1/PD2 scale-invariance control against the thresholds frozen in
\texttt{specs/PREREGISTRATION\_scale\_inv.json}.
\item[\ttfamily transformer\_sweep\_audit.py, compute\_unembedding\_rank.py] \emph{(Sec.~\ref{sec:transformer}).}
Run the well-powered transformer clamp and free-decay sweeps and compute the unembedding-locus effective
rank alongside the embedding.
\item[\ttfamily test\_compression\_clock\_adversarial\_v1.py] \emph{(adversarial tests).} Nine
pre-registered adversarial cases; checks each verdict and its reason (a clock verdict must be backed by a
finite order statistic). Caught a real false-confidence regression during development.
\item[\ttfamily test\_compression\_clock\_properties\_v1.py] \emph{(property tests).} Structural
invariants checked over many randomized trajectory sets rather than fixed shapes: no clock verdict on an
undefined order statistic, non-negative reported gap, censoring monotonic under truncation (less data
never yields more confidence), boundary-gate safety, and determinism.
\item[\ttfamily make\_figures\_pub.py] \emph{(figures).} Regenerates all publication figures directly from
the processed trajectories, printing verification numbers for cross-checking; no values are hard-coded.
\item[\ttfamily reproduce\_all.py] \emph{(driver).} Runs the test suite, the analyzer on each bundled task,
the sensitivity grid, the architecture-generality and scale-invariance audits, and the transformer
dose-response, then rebuilds every figure from processed trajectories.
\item[\ttfamily sample\_data/] \emph{(data).} Processed long-train trajectories for modular addition,
modular multiplication, and an un-normalized transformer, sufficient to reproduce the tables and figures
from analysis (no raw training required).
\end{description}

\end{document}